\documentclass[10pt,twocolumn,letterpaper]{article}
\usepackage[table]{xcolor}
\usepackage{cvpr}
\usepackage{times}
\usepackage{epsfig}
\usepackage{graphicx}
 \usepackage{multirow}
\usepackage{color}
\usepackage{cite}
\usepackage{dsfont}
\usepackage{ragged2e}
\usepackage{nccmath}
\usepackage{textcomp}
\usepackage{arydshln}
\usepackage{caption}

\usepackage[pagebackref=true,breaklinks=true,letterpaper=true,colorlinks,bookmarks=false]{hyperref}

\cvprfinalcopy 

\def\cvprPaperID{105} 

\ifcvprfinal\fi
\begin{document}

\title{Deep White-Balance Editing}

\author{Mahmoud Afifi$^\texttt{1,2}$ \hspace{1cm} Michael S. Brown$^\texttt{1}$\\
$^\texttt{1}$Samsung AI Center (SAIC) -- Toronto \hspace{0.4cm}$^\texttt{2}$York University\\
{\tt\small \{mafifi, mbrown\}@eecs.yorku.ca}}

\twocolumn[{%
\renewcommand\twocolumn[1][]{#1}%
\maketitle
\begin{center}
    \centering
    \includegraphics[width=\textwidth]{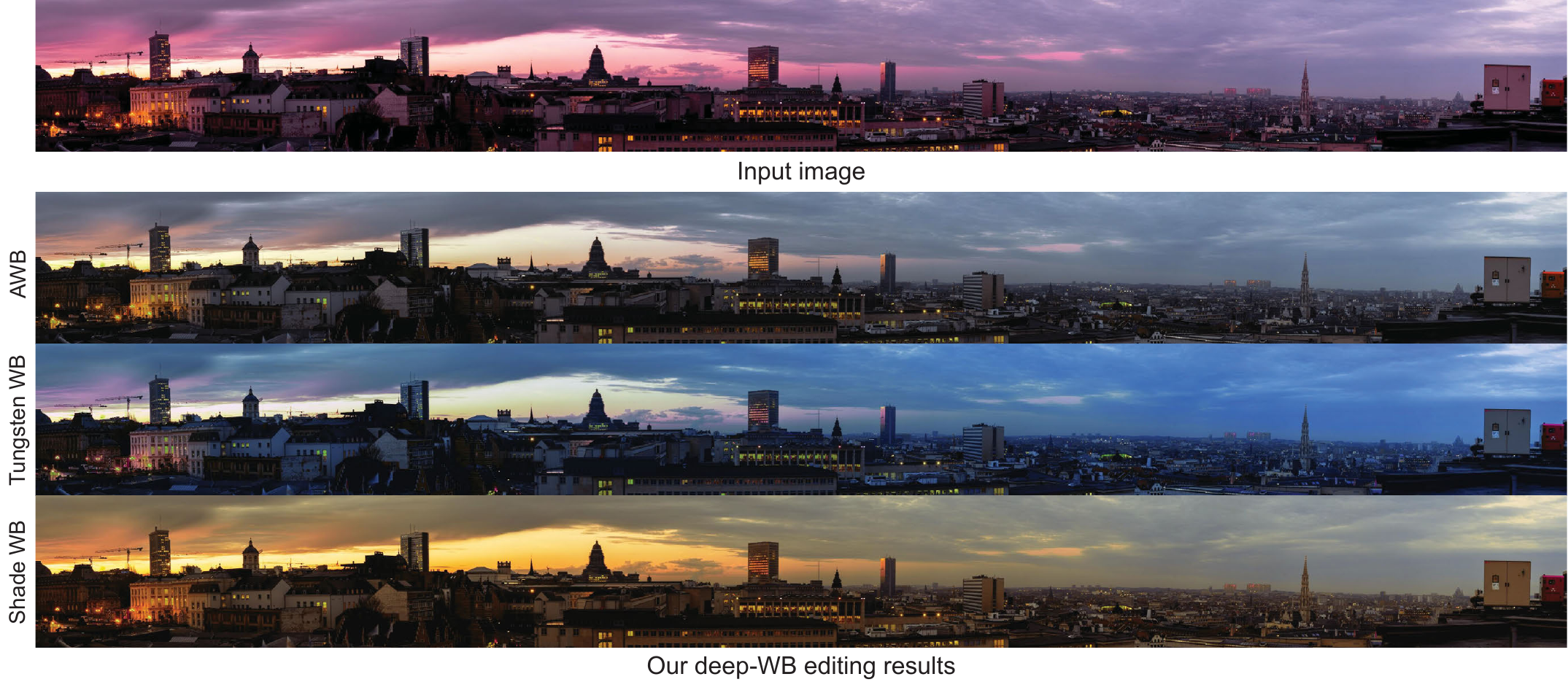}
    \captionof{figure}{Our deep white-balance editing framework produces compelling results and generalizes well to images outside our training data (e.g., image above taken from an Internet photo repository). Top: input image captured with a wrong WB setting. Bottom: our framework's AWB, Incandescent WB, and Shade WB results. Photo credit: \textit{M@tth1eu} Flickr--CC BY-NC 2.0.} \label{fig:teaser}
\end{center}%
}]

\maketitle

\begin{abstract}
We introduce a deep learning approach to realistically edit an sRGB image's white balance. Cameras capture sensor images that are rendered by their integrated signal processor (ISP) to a standard RGB (sRGB) color space encoding. The ISP rendering begins with a white-balance procedure that is used to remove the color cast of the scene's illumination.  The ISP then applies a series of nonlinear color manipulations to enhance the visual quality of the final sRGB image.  Recent work by~\cite{afifi2019color} showed that sRGB images that were rendered with the incorrect white balance cannot be easily corrected due to the ISP's nonlinear rendering.  The work in~\cite{afifi2019color} proposed a k-nearest neighbor (KNN) solution based on tens of thousands of image pairs.   We propose to solve this problem with a deep neural network (DNN) architecture trained in an end-to-end manner to learn the correct white balance.  Our DNN maps an input image to two additional white-balance settings corresponding to indoor and outdoor illuminations.  Our solution not only is more accurate than the KNN approach in terms of correcting a wrong white-balance setting but also provides the user the freedom to edit the white balance in the sRGB image to other illumination settings.
\end{abstract}

\section{Introduction and related work}\label{sec:intro_related}

White balance (WB) is a fundamental low-level computer vision task applied to all camera images. WB is performed to ensure that scene objects appear as the same color even when imaged under different illumination conditions. Conceptually,  WB is intended to normalize the effect of the captured scene's illumination such that all objects appear as if they were captured under ideal ``white light''.  WB is one of the first color manipulation steps applied to the sensor's unprocessed raw-RGB image by the camera's onboard integrated signal processor (ISP).  After WB is performed, a number of additional color rendering steps are applied by the ISP to further process the raw-RGB image to its final standard RGB (sRGB) encoding.  While the goal of WB is intended to normalize the effect of the scene's illumination, ISPs often incorporate aesthetic considerations in their color rendering based on photographic preferences. Such preferences do not always conform to the white light assumption and can vary based on different factors, such as cultural preference and scene content~\cite{scuello2004museum, cheng2016two, barron2017fast, hu2018exposure}.

Most digital cameras provide an option to adjust the WB settings during image capturing. However, once the WB setting has been selected and the image is fully processed by the ISP to its final sRGB encoding it becomes challenging to perform WB editing without access to the original unprocessed raw-RGB image~\cite{afifi2019color}. This problem becomes even more difficult if the WB setting was wrong, which results in a strong color cast in the final sRGB image.

The ability to edit the WB of an sRGB image not only is useful from a photographic perspective but also can be beneficial for computer vision applications, such as object recognition, scene understanding, and color augmentation~\cite{barnard2002comparison, gijsenij2011computational, afifi2019ICCV}. A recent study in \cite{afifi2019ICCV} showed that images captured with an incorrect WB setting produce a similar effect of an untargeted adversarial attack for deep neural network (DNN) models.

\paragraph{In-camera WB procedure}  To understand the challenge of WB editing in sRGB images it is useful to review how cameras perform WB.  WB consists of two steps performed in tandem by the ISP: (1) estimate the camera sensor's response to the scene illumination in the form of a raw-RGB vector; (2) divide each R/G/B color channel in the raw-RGB image by the corresponding  channel response in the raw-RGB vector.   The first step of estimating the illumination vector constitutes the camera's auto-white-balance (AWB) procedure.   Illumination estimation is a well-studied topic in computer vision---representative works include \cite{buchsbaum1980spatial, SoG, cheng2014illuminant, gehler2008bayesian, barron2015convolutional, shi2016deep, barron2017fast, hu2017fc, bianco2019quasi, qian2019finding, afifi2019SIIE}. In addition to AWB, most cameras allow the user to manually select among WB presets in which the raw-RGB vector for each preset has been determined by the camera manufacturer. These presets correspond to common scene illuminants (e.g., Daylight, Shade, Incandescent).

Once the scene's illumation raw-RGB vector is defined, a simple linear scaling is applied to each color channel independently to normalize the illumination. This scaling operation is performed using a $3\!\times\!3$ diagonal matrix. The white-balanced raw-RGB image is then further processed by camera-specific ISP steps, many nonlinear in nature, to render the final images in an output-referred color space---namely, the sRGB color space. These nonlinear operations make it hard to use the traditional diagonal correction to correct images rendered with strong color casts caused by camera WB errors \cite{afifi2019color}.

\paragraph{WB editing in sRGB} In order to perform accurate post-capture WB editing, the rendered sRGB values should be properly reversed to obtain the corresponding unprocessed raw-RGB values and then re-rendered. This can be achieved only by  accurate radiometric calibration methods (e.g., \cite{kim2012new, colorderenderingCVPR2012, chakrabarti2014modeling}) that compute the necessary metadata for such color de-rendering.   Recent work by Afifi et al.~\cite{afifi2019color} proposed a method to directly correct sRGB images that were captured with the wrong WB setting. This work proposed an exemplar-based framework using a large dataset of over 65,000 sRGB images rendered by a software camera pipeline with the wrong WB setting.  Each of these sRGB images had a corresponding sRGB image that was rendered with the correct WB setting.  Given an input image, their approach used a KNN strategy to find similar images in their dataset and computed a mapping function to the corresponding correct WB images.  The work in~\cite{afifi2019color} showed that this computed color mapping constructed from exemplars was effective in correcting an input image. Later Afifi and Brown~\cite{afifi2019ICCV} extended their KNN idea to map a correct WB image to appear incorrect for the purpose of image augmentation for training deep neural networks.  Our work is inspired by~\cite{afifi2019color, afifi2019ICCV} in their effort to directly edit the WB in an sRGB image. However, in contrast to the KNN frameworks in~\cite{afifi2019color, afifi2019ICCV}, we cast the problem within a single deep learning framework that can achieve both tasks---namely, WB correction and WB manipulation as shown in Fig.\ \ref{fig:teaser}.

\paragraph{Contribution}~We present a novel deep learning framework that allows realistic post-capture WB editing of sRGB images. Our framework consists of a single encoder network that is coupled with three decoders targeting the following WB settings: (1) a ``correct'' AWB setting; (2) an indoor WB setting; (3) an outdoor WB setting.   The first decoder allows an sRGB image that has been incorrectly white-balanced image to be edited to have the correct WB.  This is useful for the task of post-capture WB correction.  The additional indoor and outdoor decoders provide users the ability to produce a wide range of different WB appearances by blending between the two outputs.  This supports photographic editing tasks to adjust an image's aesthetic WB properties.  We provide extensive experiments to demonstrate that our method generalizes well to images outside our training data and achieves state-of-the-art results for both tasks.

\begin{figure*}[!t]
\centering
\includegraphics[width=1\linewidth]{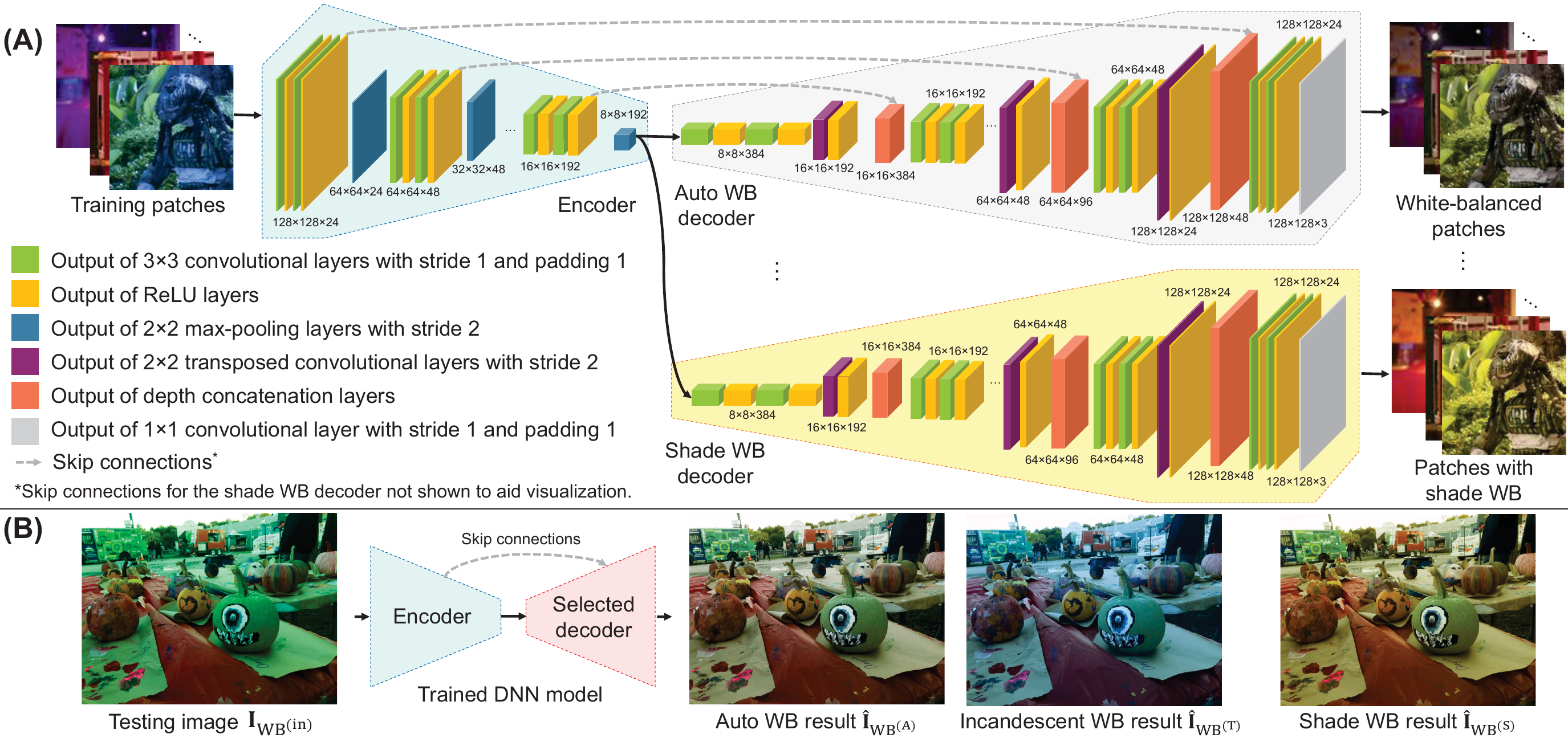}
\caption{Proposed multi-decoder framework for sRGB WB editing. (A) Our proposed framework consists of a single encoder and multiple decoders. The training process is performed in an end-to-end manner, such that each decoder ``re-renders'' the given training patch with a specific WB setting, including AWB. For training, we randomly select image patches from the Rendered WB dataset~\cite{afifi2019color}. (B) Given a testing image, we produce the targeted WB setting by using the corresponding trained decoder. }
\label{fig:main}
\end{figure*}

\section{Deep white-balance editing}\label{sec:method}

\subsection{Problem formulation} \label{subsec:formulation}
Given an sRGB image, $\mathbf{I}_{\text{WB}^{(\text{in})}}$, rendered through an unknown camera ISP with an arbitrary WB setting $\text{WB}^{(\text{in})}$, our goal is to edit its colors to appear as if it were re-rendered with a target WB setting $\text{WB}^{(t)}$.

As mentioned in Sec.~\ref{sec:intro_related}, our task can be accomplished accurately if the original unprocessed raw-RGB image is available. If we could recover the unprocessed raw-RGB values, we can change the WB setting $\text{WB}^{(\text{in})}$ to $\text{WB}^{(t)}$, and then re-render the image back to the sRGB color space with a software-based ISP. This ideal process can be described by the following equation:
\begin{equation}
\label{eq1}
\mathbf{I}_{\text{WB}^{(t)}} =  G\left(F  \left(\mathbf{I}_{\text{WB}^{(\text{in})}}\right)\right),
\end{equation}

\noindent where $F: \mathbf{I}_{\text{WB}^{(\text{in})}} \rightarrow  \mathbf{D}_{\text{WB}^{(\text{in})}}$ is an unknown reconstruction function that reverses the camera-rendered sRGB image $\mathbf{I}$ back to its corresponding raw-RGB image $\mathbf{D}$ with the current $\text{WB}^{(\text{in})}$ setting applied and $G: \mathbf{D}_{\text{WB}^{(\text{in})}} \rightarrow  \mathbf{I}_{\text{WB}^{(t)}}$ is an unknown camera rendering function that is responsible for editing the WB setting and re-rendering the final image.

\subsection{Method overview} \label{subsec:overview}

\begin{figure*}
\centering
\includegraphics[width=\linewidth]{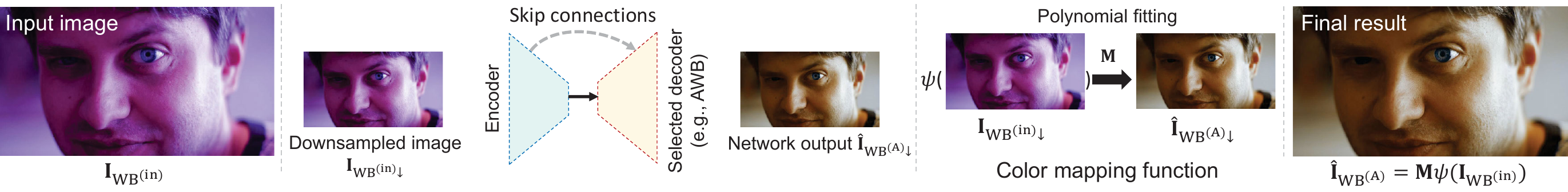}
\caption{We consider the runtime performance of our method to be able to run on limited computing resources ($\sim$1.5 seconds on a single CPU to process a 12-megapixel image). First, our DNN processes a downsampled version of the input image, and then we apply a global color mapping to produce the output image in its original resolution. The shown input image is rendered from the MIT-Adobe FiveK dataset \cite{fivek}.}\vspace{-4mm}
\label{fig:entier_process}
\end{figure*}

Our goal is to model the functionality of $G\left(F\left(\cdot\right)\right)$ to generate $\mathbf{I}_{\text{WB}^{(t)}}$. We first analyze how the functions $G$ and $F$ cooperate to produce $\mathbf{I}_{\text{WB}^{(t)}}$. From Eq.\ \ref{eq1}, we see that the function $F$ transforms the input image $\mathbf{I}_{\text{WB}^{(\text{in})}}$ into an intermediate representation (i.e., the raw-RGB image with the captured WB setting), while the function $G$ accepts this intermediate representation and renders it with the target WB setting to an sRGB color space encoding.

Due to the nonlinearities applied by the ISP's rendering chain, we can think of $G$ as a hybrid function that consists of a set of sub-functions, where each sub-function is responsible for rendering the intermediate representation with a specific WB setting.

Our ultimate goal is \textit{not} to reconstruct/re-render the original raw-RGB values, but rather to generate the final sRGB image with the target WB setting ${\text{WB}^{(t)}}$. Therefore, we can model the functionality of $G\left(F\left(\cdot\right)\right)$ as an encoder/decoder scheme. Our encoder $f$ transfers the input image into a latent representation, while each of our decoders ($g_1$, $g_2$, ...) generates the final images with a different WB setting. Similar to Eq.\ \ref{eq1}, we can formulate our framework as follows:

\begin{equation}
\label{eq2}
\hat{\mathbf{I}}_{\text{WB}^{(\text{t})}} = g_t\left(f \left( \mathbf{I}_{\text{WB}^{(\text{in})}} \right)\right),
\end{equation}

\noindent where $f: \mathbf{I}_{\text{WB}^{(\text{in})}} \rightarrow \mathbf{\mathcal{Z}}$, $g_t: \mathbf{\mathcal{Z}} \rightarrow \hat{\mathbf{I}}_{\text{WB}^{(\text{t})}}$, and $\mathcal{Z}$ is an intermediate representation (i.e., latent representation) of the original input image $\mathbf{I}_{\text{WB}^{(\text{in})}}$.

Our goal is to make the functions $f$ and $g_t$ independent, such that changing $g_t$ with a new function $g_y$ that targets a different WB $y$ does not require any modification in $f$, as is the case in Eq.\ \ref{eq1}.

In our work, we target three different WB settings: (i) $\text{WB}^{(\text{A})}$: AWB---representing the correct lighting of the captured image's scene; (ii) $\text{WB}^{(\text{T})}$: Tungsten/Incandescent---representing WB for indoor lighting; and (iii) $\text{WB}^{(\text{S})}$: Shade---representing WB for outdoor lighting.  This gives rise to three different decoders ($g_A$, $g_T$, and $g_S$) that are responsible for generating output images that correspond to AWB, Incandescent WB, and Shade WB.

The Incandescent and Shade WB are specifically selected based on the color properties.  This can be understood when considering the illuminations in terms of their correlated color temperatures. For example, Incandescent and Shade WB settings are correlated to $2850$ Kelvin ($K$) and $7500K$ color temperatures, respectively. This wide range of illumination color temperatures considers the range of pleasing illuminations \cite{kruithof1941tubular, petrulis2018exploring}. Moreover, the wide color temperature range between Incandescent and Shade allows the approximation of images with color temperatures within this range by interpolation. The details of this interpolation process are explained in Sec. \ref{subsec:testing_phase}. Note that there is no fixed correlated color temperature for the AWB mode, as it changes based on the input image's lighting conditions.

\subsection{Multi-decoder architecture} \label{subsec:architecture}

An overview of our DNN's architecture is shown in Fig.\ \ref{fig:main}. We use a U-Net architecture \cite{unet} with multi-scale skip connections between the encoder and decoders. Our framework consists of two main units:  the first is a 4-level encoder unit that is responsible for extracting a multi-scale latent representation of our input image; the second unit includes three 4-level decoders.  Each unit has a different bottleneck and transposed convolutional (conv) layers. At the first level of our encoder and each decoder, the conv layers have 24 channels. For each subsequent level, the number of channels is doubled (i.e., the fourth level has 192 channels for each conv layer).

\subsection{Training phase} \label{subsec:training}

\paragraph{Training data}
We adopt the Rendered WB dataset produced by~\cite{afifi2019color} to train and validate our model. This dataset includes $\sim$65K sRGB images rendered by different camera models and with different WB settings, including the Shade and Incandescent settings. For each image, there is also a corresponding ground truth image rendered with the correct WB setting (considered to be the correct AWB result). This dataset consists of two subsets: Set 1 (62,535 images taken by seven different DSLR cameras) and Set 2 (2,881 images taken by a DSLR camera and four mobile phone cameras). The first set (i.e., Set 1) is divided into three equal partitions by \cite{afifi2019color}. We randomly selected 12,000 training images from the first two partitions of Set 1 to train our model. For each training image, we have three ground truth images rendered with: (i) the correct WB (denoted as AWB), (ii) Shade WB, and (iii) Incandescent WB. The final partition of Set 1 (21,046 images) is used for testing. We refer to this partition as Set 1--\textit{Test}. Images of Set 2 are not used in training and the entire set is used for testing.

\begin{figure}
\centering
\includegraphics[width=\linewidth]{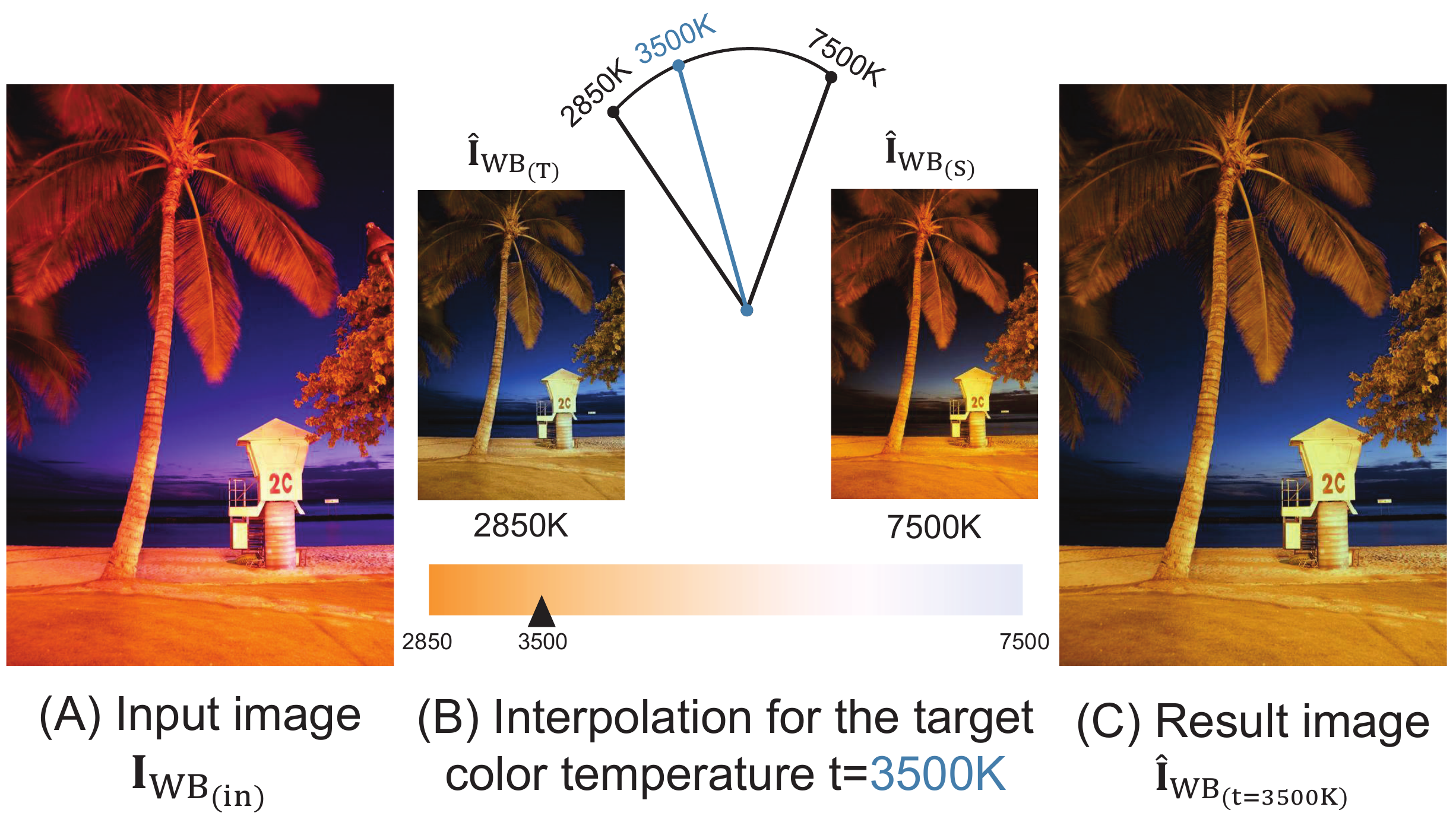}
\caption{In addition to our AWB correction, we train our framework to produce two different color temperatures (i.e., Incandescent and Shade WB settings). We interpolate between these settings to produce images with other color temperatures. (A) Input image. (B) Interpolation process. (C) Final result. The shown input image is taken from the rendered version of the MIT-Adobe FiveK dataset \cite{fivek, afifi2019color}.}
\label{fig:tempblending}
\end{figure}

\paragraph{Data augmentation} We also augment the training images by rendering an additional 1,029 raw-RGB images, of the same scenes included in the Rendered WB dataset \cite{afifi2019color}, but with random color temperatures. At each epoch, we randomly select four $128\!\times\!128$ patches from each training image and their corresponding ground truth images for each decoder and apply geometric augmentation (rotation and flipping) as an additional data augmentation to avoid overfitting.

\paragraph{Loss function}
We trained our model to minimize the $\texttt{L}_1$-norm loss function between the reconstructed and ground truth patches:

\begin{equation}
\label{eq3} \sum_{i}\sum_{p=1}^{3hw}\left|\mathbf{P}_{\text{WB}^{(i)}}(p) - \mathbf{C}_{\text{WB}^{(i)}}(p)\right|,
\end{equation}

\noindent where $h$ and $w$ denote the patch's height and width, and $p$ indexes into each pixel of the training patch $\mathbf{P}$ and the ground truth camera-rendered patch $\mathbf{C}$, respectively. The index $i\in\{\text{A},\text{T},\text{S}\}$ refers to the three target WB settings. We also have examined the squared $\texttt{L}_2$-norm loss function and found that both loss functions work well for our task.

\begin{figure*}[t]
\centering
\includegraphics[width=\linewidth]{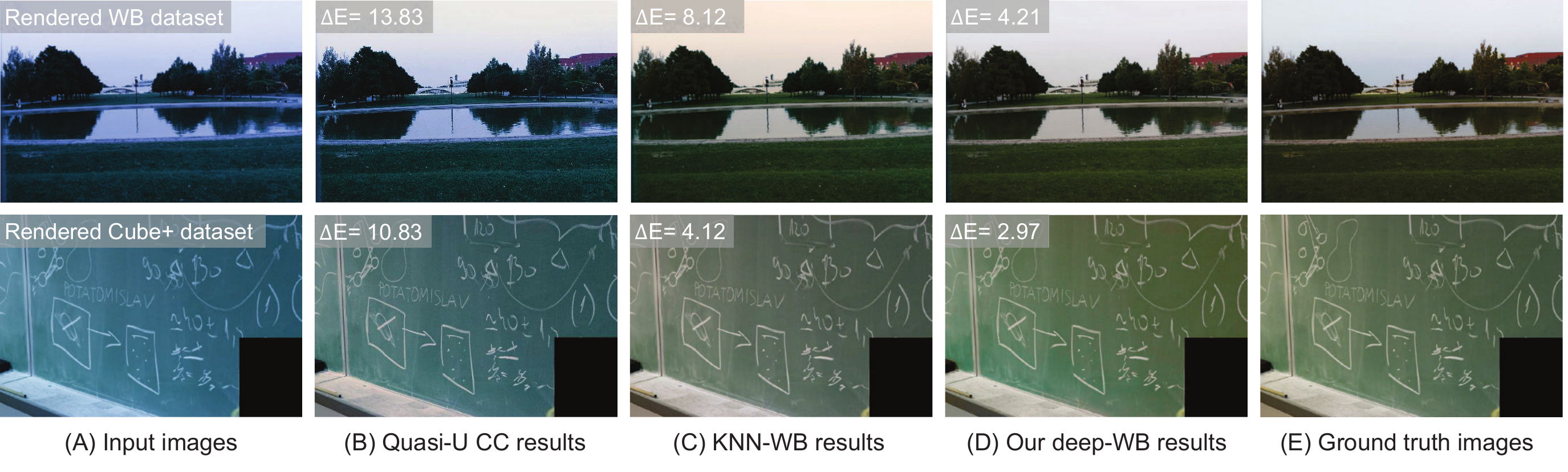}
\caption{Qualitative comparison of AWB correction. (A) Input images. (B) Results of quasi-U CC \cite{bianco2019quasi}. (C) Results of KNN-WB \cite{afifi2019color}. (D) Our results. (E) Ground truth images. Shown input images are taken from the Rendered WB dataset \cite{afifi2019color} and the rendered version of Cube+ dataset \cite{banic2017unsupervised, afifi2019color}.}
\label{fig:qualitative_AWB}
\end{figure*}

\paragraph{Training details}
We initialized the weights of the conv layers using He's initialization \cite{he2015delving}. The training process is performed for 165,000 iterations using the adaptive moment estimation (Adam) optimizer \cite{kingma2014adam}, with a decay rate of gradient moving average $\beta_1=0.9$ and a decay rate of squared gradient moving average $\beta_2=0.999$. We used a learning rate of $10^{-4}$ and reduced it by 0.5 every 25 epochs. The mini-batch size was 32 training patches per iteration. Each mini-batch contains random patches selected from training images that may contain different WB settings. During training, each decoder receives the generated latent representations by our single encoder and generates corresponding patches with the target WB setting. The loss function is computed using the result of each decoder and is followed by gradient backpropagation from all decoders aggregated back to our single encoder via the skip-layer connections. Thus, the encoder is trained to map the images into an intermediate latent space that is beneficial for generating the target WB setting by each decoder.

\subsection{Testing phase} \label{subsec:testing_phase}

\paragraph{Color mapping procedure}
Our DNN model is a fully convolutional network and is able to process input images in their original dimensions with the restriction that the dimensions should be multiples of $2^4$, as we use 4-level encoder/decoders with $2\!\times\!2$ max-pooling and transposed conv layers. However, to ensure a consistent run time for any sized input images, we resize all input images to a maximum dimension of 656 pixels. Our DNN is applied on this resized image to produce image $\hat{\mathbf{I}}_{\text{WB}^{(i)}\downarrow}$ with the target WB setting $i\in\{\text{A},\text{T},\text{S}\}$.

We then compute a color mapping function between our resized input and output image. Work in~\cite{finlayson2015color, hong2001study} evaluated several types of polynomial mapping functions and showed their effectiveness to achieve nonlinear color mapping. Accordingly, we computed a polynomial mapping matrix $\mathbf{M}$ that globally maps values of $\psi\left(\mathbf{I}_{\text{WB}^{(\text{in})}\downarrow}\right)$ to the colors of our generated image $\hat{\mathbf{I}}_{\text{WB}^{(i)}\downarrow}$, where $\psi(\cdot)$ is a polynomial kernel function that maps the image's RGB vectors to a higher 11-dimensional space. This mapping matrix $\mathbf{M}$ can be computed in a closed-form solution, as demonstrated in \cite{afifi2019color, afifi2019ICCV}.

Once $\mathbf{M}$ is computed, we obtain our final result in the same input image resolution using the following equation \cite{afifi2019color}:

\begin{equation}
\label{eq5}
\hat{\mathbf{I}}_{\text{WB}^{(\text{i})}} =  \mathbf{M}\psi\left(\mathbf{I}_{\text{WB}^{(\text{in})}}\right).
\end{equation}

Fig.\ \ref{fig:entier_process} illustrates our color mapping procedure. Our method requires $\sim$1.5 seconds on an Intel Xeon E5-1607 @ 3.10GHz machine with 32 GB RAM to process a 12-megapixel image for a selected WB setting.

We note that an alternative strategy is to compute the color polynomial mapping matrix directly~\cite{schwartz2018deepisp}.  We conducted preliminary experiments and found that estimating the polynomial matrix directly was less robust than generating the image itself followed by fitting a global polynomial function. The reason is that having small errors in the estimated polynomial coefficients can lead to noticeable color errors (e.g., out-of-gamut values), whereas small errors in the estimated image were ameliorated by the global fitting.

\paragraph{Editing by user manipulation}

Our framework allows the user to choose between generating result images with the three available WB settings (i.e., AWB, Shade WB, and Incandescent WB).  Using the Shade and Incandescent WB, the user can edit the image to a specific WB setting in terms of color temperature, as explained in the following.

To produce the effect of a \textit{new} target WB setting with a color temperature $t$ that is not produced by our decoders, we can interpolate between our generated images with the Incandescent and Shade WB settings. We found that a simple linear interpolation was sufficient for this purpose. This operation is described by the following equation:

\begin{equation}
\label{eq6}
\hat{\mathbf{I}}_{\text{WB}^{(t)}} = b \text{ }\hat{\mathbf{I}}_{\text{WB}^{(\text{T})}} + (1-b) \text{ } \hat{\mathbf{I}}_{\text{WB}^{(\text{S})}},
\end{equation}

\noindent where $\hat{\mathbf{I}}_{\text{WB}^{(\text{T})}}$ and $\hat{\mathbf{I}}_{\text{WB}^{(\text{S})}}$ are our produced images with Incandescent and Shade WB settings, respectively, and $b$ is the interpolation ratio that is given by $\frac{1/t - 1/t(S)}{1/t(T) - 1/t(S)}$. Fig.\ \ref{fig:tempblending} shows an example.

\begin{table*}[]
\centering
\caption{AWB results using the Rendered WB dataset \cite{afifi2019color} and the rendered version of the Cube+ dataset \cite{banic2017unsupervised, afifi2019color}. We report the mean, first, second (median), and third quartile (Q1, Q2, and Q3) of mean square error (MSE), mean angular error (MAE), and $\boldsymbol{\bigtriangleup}$\textbf{E} 2000 \cite{sharma2005ciede2000}. For all diagonal-based methods, gamma linearization \cite{anderson1996proposal, ebner2007color} is applied. The top results are indicated with yellow and boldface.}
\label{Table0}
\scalebox{0.9}{
\begin{tabular}{|l|c|c|c|c|c|c|c|c|c|c|c|c|}
\hline
\multicolumn{1}{|c|}{} & \multicolumn{4}{c|}{\textbf{MSE}} & \multicolumn{4}{c|}{\textbf{MAE}} & \multicolumn{4}{c|}{\textbf{$\boldsymbol{\bigtriangleup}$\textbf{E} 2000}} \\ \cline{2-13}
\multicolumn{1}{|c|}{\multirow{-2}{*}{\textbf{Method}}} & \textbf{Mean} & \textbf{Q1} & \textbf{Q2} & \textbf{Q3} & \textbf{Mean} & \textbf{Q1} & \textbf{Q2} & \textbf{Q3} & \textbf{Mean} & \textbf{Q1} & \textbf{Q2} & \textbf{Q3}  \\ \hline

\multicolumn{13}{|c|}{\cellcolor[HTML]{D4EBF2}\textbf{Rendered WB dataset: Set 1--\textit{Test} (21,046 images) \cite{afifi2019color}}} \\ \hline

FC4 \cite{hu2017fc} & 179.55 & 33.89 & 100.09 & 246.50 & 6.14\textdegree & 2.62\textdegree & 4.73\textdegree & 8.40\textdegree &  6.55 & 3.54 & 5.90 & 8.94  \\\hline

Quasi-U CC \cite{bianco2019quasi}   & 172.43 & 33.53 & 97.9 & 237.26 & 6.00\textdegree & 2.79\textdegree & 4.85\textdegree & 8.15\textdegree &	6.04 & 3.24 & 5.27 & 8.11 \\ \hline

KNN-WB \cite{afifi2019color} &  \cellcolor[HTML]{FFFFBB}\textbf{77.79} &  13.74& \cellcolor[HTML]{FFFFBB} \textbf{39.62}& \cellcolor[HTML]{FFFFBB}\textbf{94.01 }&  \cellcolor[HTML]{FFFFBB} \textbf{3.06}\textdegree & \cellcolor[HTML]{FFFFBB} \textbf{1.74}\textdegree & \cellcolor[HTML]{FFFFBB} \textbf{2.54}\textdegree &  \cellcolor[HTML]{FFFFBB}\textbf{3.76}\textdegree &  \cellcolor[HTML]{FFFFBB}\textbf{3.58} &  \cellcolor[HTML]{FFFFBB} \textbf{2.07} &  \cellcolor[HTML]{FFFFBB} \textbf{3.09} &  \cellcolor[HTML]{FFFFBB} \textbf{4.55} \\ \hdashline

Ours & 82.55 & \cellcolor[HTML]{FFFFBB} \textbf{13.19} & 42.77 & 102.09 & 3.12\textdegree & 1.88\textdegree & 2.70\textdegree & 3.84\textdegree & 3.77 & 2.16 & 3.30 & 4.86 \\ \hline

\multicolumn{13}{|c|}{\cellcolor[HTML]{D4EBF2}\textbf{Rendered WB dataset: Set 2 (2,881 images) \cite{afifi2019color}}} \\ \hline

FC4 \cite{hu2017fc}  &  505.30 & 142.46 & 307.77 & 635.35 & 10.37\textdegree
& 5.31\textdegree & 9.26\textdegree & 14.15\textdegree & 10.82 & 7.39 & 10.64 &  13.77   \\ \hline

Quasi-U CC \cite{bianco2019quasi} & 553.54 & 146.85 & 332.42 & 717.61 & 10.47\textdegree & 5.94\textdegree & 9.42\textdegree & 14.04\textdegree & 10.66 & 7.03 & 10.52 & 13.94 \\ \hline

KNN-WB \cite{afifi2019color} & 171.09 & 37.04 & 87.04 & 190.88 & 4.48\textdegree &  2.26\textdegree & 3.64\textdegree & 5.95\textdegree & 5.60 & 3.43 &  4.90 & 7.06  \\ \hdashline

Ours & \cellcolor[HTML]{FFFFBB} \textbf{124.97} &
\cellcolor[HTML]{FFFFBB} \textbf{30.13} &
\cellcolor[HTML]{FFFFBB} \textbf{76.32} &
\cellcolor[HTML]{FFFFBB} \textbf{154.44} &
\cellcolor[HTML]{FFFFBB} \textbf{3.75}\textdegree &
\cellcolor[HTML]{FFFFBB} \textbf{2.02}\textdegree &
\cellcolor[HTML]{FFFFBB} \textbf{3.08}\textdegree &
\cellcolor[HTML]{FFFFBB} \textbf{4.72}\textdegree &
\cellcolor[HTML]{FFFFBB} \textbf{4.90} &
\cellcolor[HTML]{FFFFBB} \textbf{3.13} &
\cellcolor[HTML]{FFFFBB} \textbf{4.35}  &
\cellcolor[HTML]{FFFFBB} \textbf{6.08} \\ \hline

\multicolumn{13}{|c|}{\cellcolor[HTML]{D4EBF2}\textbf{Rendered Cube+ dataset with different WB settings (10,242 images) \cite{banic2017unsupervised, afifi2019color}}} \\ \hline

FC4 \cite{hu2017fc}  & 371.9 & 79.15 & 213.41 & 467.33 & 6.49\textdegree & 3.34\textdegree & 5.59\textdegree & 8.59\textdegree & 10.38 & 6.6 & 9.76 & 13.26 \\ \hline

Quasi-U CC \cite{bianco2019quasi} & 292.18& 15.57 & 55.41 & 261.58 & 6.12\textdegree & 1.95\textdegree & 3.88\textdegree & 8.83\textdegree & 7.25 & 2.89 & 5.21 & 10.37 \\ \hline

KNN-WB \cite{afifi2019color} &  194.98 &  27.43 &  57.08 &  118.21 &  4.12\textdegree &  1.96\textdegree & 3.17\textdegree & 5.04\textdegree &  5.68 &  3.22 &  4.61 &  6.70
\\ \hdashline

Ours & \cellcolor[HTML]{FFFFBB} \textbf{80.46} &
\cellcolor[HTML]{FFFFBB} \textbf{15.43} &
\cellcolor[HTML]{FFFFBB} \textbf{33.88} &
\cellcolor[HTML]{FFFFBB} \textbf{74.42} &
\cellcolor[HTML]{FFFFBB} \textbf{3.45}\textdegree &
\cellcolor[HTML]{FFFFBB} \textbf{1.87}\textdegree &
\cellcolor[HTML]{FFFFBB} \textbf{2.82}\textdegree &
\cellcolor[HTML]{FFFFBB} \textbf{4.26}\textdegree &
\cellcolor[HTML]{FFFFBB} \textbf{4.59} &
\cellcolor[HTML]{FFFFBB} \textbf{2.68} &
\cellcolor[HTML]{FFFFBB} \textbf{3.81}  &
\cellcolor[HTML]{FFFFBB} \textbf{5.53} \\ \hline

\end{tabular}
}\vspace{2mm}
\end{table*}

\section{Results}\label{sec:results}

Our method targets two different tasks: post-capture WB correction and manipulation of the sRGB rendered images to a specific WB color temperature. We achieve state-of-the-art results for several different datasets for both tasks. We first describe the datasets used to evaluate our method in Sec.~\ref{subsec:datasets}. We then discuss our quantitative and qualitative results in Sec.~\ref{subsec:quantitative} and Sec. \ref{subsec:qualitative}, respectively. We also perform an ablation study to validate our problem formulation and the proposed framework.

\begin{figure}[t]
\centering

\includegraphics[width=\linewidth]{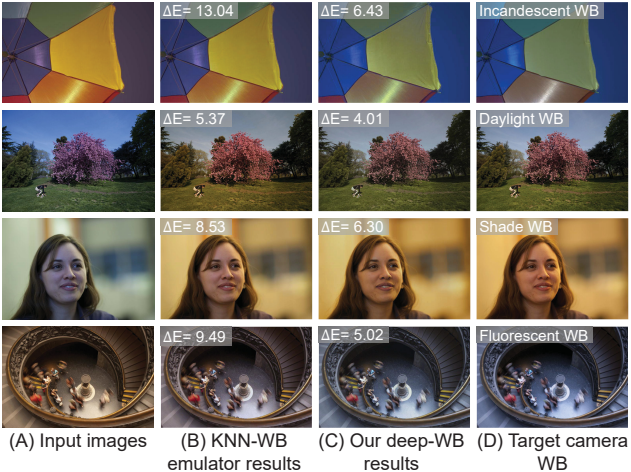}
\caption{Qualitative comparison of WB manipulation. (A) Input images. (B) Results of KNN-WB emulator \cite{afifi2019ICCV}. (C) Our results. (D) Ground truth camera-rendered images with the target WB settings. In this figure, the target WB settings are Incandescent, Daylight, Shade, and Fluorescent. Shown input images are taken from the rendered version of the MIT-Adobe FiveK dataset \cite{fivek, afifi2019color}.}
\label{fig:qualitative_WBEditing}
\end{figure}

\subsection{Datasets}\label{subsec:datasets}
As previously mentioned, we used randomly selected images from the two partitions of Set 1 in the Rendered WB dataset \cite{afifi2019color} for training. For testing, we used the third partition of Set 1, termed Set 1-\textit{Test}, and three additional datasets not part of training.  Two of these additional datasets are as follows: (1) Set 2 of the Rendered WB dataset (2,881 images) \cite{afifi2019color}, and (2) the sRGB rendered version of the Cube+ dataset (10,242 images) \cite{banic2017unsupervised}. Datasets (1) and (2) are used to evaluate the task of AWB correction. For the WB manipulation task, we used the rendered Cube+ dataset and (3) a rendered version of the MIT-Adobe FiveK dataset (29,980 images) \cite{fivek}. The rendered version of each dataset of these datasets is available from the project page associated with \cite{afifi2019color}.  These latter datasets represent raw-RGB images that have been rendered to the sRGB color space with {\it different} WB settings. This allows us to evaluate how well we can mimic different WB settings.

\subsection{Quantitative results}\label{subsec:quantitative}

For both tasks, we follow the same evaluation metrics used by the most recent work in \cite{afifi2019color}. Specifically, we used the following metrics to evaluate our results: mean square error (MSE), mean angular error (MAE), and $\bigtriangleup$E 2000 \cite{sharma2005ciede2000}. For each evaluation metric, we report the mean, lower quartile (Q1), median (Q2), and the upper quartile (Q3) of the error.

\begin{figure*}[]
\centering
\includegraphics[width=\linewidth]{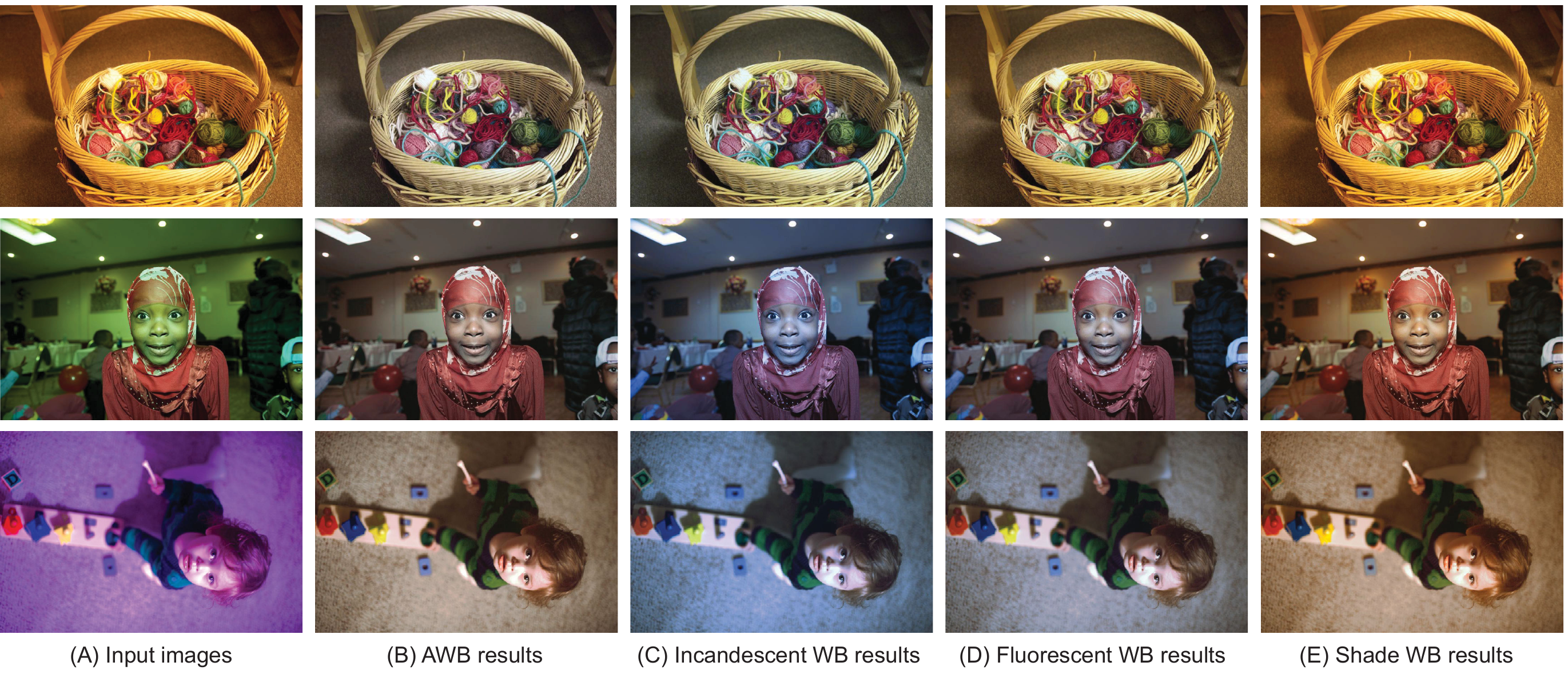}
\caption{Qualitative results of our method. (A) Input images. (B) AWB results. (C) Incandescent WB results. (D) Fluorescent WB results. (E) Shade WB Results. Shown input images are rendered from the MIT-Adobe FiveK dataset \cite{fivek}. }\vspace{-2mm}
\label{fig:qualitative}
\end{figure*}

\begin{figure*}[t]
\centering
\includegraphics[width=\linewidth]{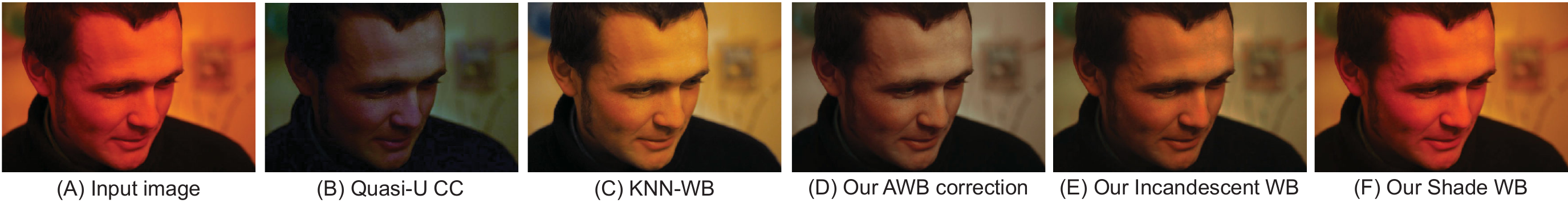}
\caption{(A) Input image. (B) Result of quasi-U CC \cite{bianco2019quasi}. (C) Result of KNN-WB \cite{afifi2019color}. (D)-(F) Our deep-WB editing results. Photo credit: \textit{Duncan Yoyos} Flickr--CC BY-NC 2.0.}
\label{fig:hard_case}
\end{figure*}

\paragraph{WB correction}  We compared the proposed method with the KNN-WB approach in \cite{afifi2019color}. We also compared our results against the traditional WB diagonal-correction using recent illuminant estimation methods~\cite{bianco2019quasi, hu2017fc}. We note that methods \cite{bianco2019quasi, hu2017fc} were not designed to correct nonlinear sRGB images. These methods are included, because it is often purported that such methods are effective when the sRGB image has been ``linearized'' using a decoding gamma.


\begin{table*}[t]
\centering
\caption{Results of WB manipulation using the rendered version of the Cube+ dataset \cite{banic2017unsupervised, afifi2019color} and the rendered version of the MIT-Adobe FiveK dataset \cite{fivek, afifi2019color}. We report the mean, first, second (median), and third quartile (Q1, Q2, and Q3) of mean square error (MSE), mean angular error (MAE), and $\boldsymbol{\bigtriangleup}$\textbf{E} 2000 \cite{sharma2005ciede2000}. The top results are indicated with yellow and boldface.}
\label{Table1}
\scalebox{0.89}{
\begin{tabular}{|l|c|c|c|c|c|c|c|c|c|c|c|c|}
\hline
\multicolumn{1}{|c|}{} & \multicolumn{4}{c|}{\textbf{MSE}} & \multicolumn{4}{c|}{\textbf{MAE}} & \multicolumn{4}{c|}{\textbf{$\boldsymbol{\bigtriangleup}$\textbf{E} 2000}} \\ \cline{2-13}
\multicolumn{1}{|c|}{\multirow{-2}{*}{\textbf{Method}}} & \textbf{Mean} & \textbf{Q1} & \textbf{Q2} & \textbf{Q3} & \textbf{Mean} & \textbf{Q1} & \textbf{Q2} & \textbf{Q3} & \textbf{Mean} & \textbf{Q1} & \textbf{Q2} & \textbf{Q3}  \\ \hline

\multicolumn{13}{|c|}{\cellcolor[HTML]{D4EBF2}\textbf{Rendered Cube+ dataset (10,242 images) \cite{banic2017unsupervised, afifi2019color}}} \\ \hline

KNN-WB emulator \cite{afifi2019ICCV} & 317.25 & 50.47 & 153.33 & 428.32 &  7.6\textdegree &  3.56\textdegree & 6.15\textdegree & 10.63\textdegree &  7.86 & 4.00 & 6.56 &  10.46
 \\ \hdashline
 					
Ours & \cellcolor[HTML]{FFFFBB} \textbf{199.38} &
\cellcolor[HTML]{FFFFBB} \textbf{32.30} &
\cellcolor[HTML]{FFFFBB} \textbf{63.34} &
\cellcolor[HTML]{FFFFBB} \textbf{142.76} &
\cellcolor[HTML]{FFFFBB} \textbf{5.40}\textdegree &
\cellcolor[HTML]{FFFFBB} \textbf{2.67}\textdegree &
\cellcolor[HTML]{FFFFBB} \textbf{4.04}\textdegree &
\cellcolor[HTML]{FFFFBB} \textbf{6.36}\textdegree &
\cellcolor[HTML]{FFFFBB} \textbf{5.98} &
\cellcolor[HTML]{FFFFBB} \textbf{3.44} &
\cellcolor[HTML]{FFFFBB} \textbf{4.78}  &
\cellcolor[HTML]{FFFFBB} \textbf{7.29} \\ \hline

\multicolumn{13}{|c|}{\cellcolor[HTML]{D4EBF2}\textbf{Rendered MIT-Adobe FiveK dataset (29,980 images) \cite{fivek, afifi2019color}}} \\ \hline

KNN-WB emulator \cite{afifi2019ICCV} & 249.95 & 41.79 & 109.69 & 283.42 & 7.46\textdegree &  3.71\textdegree & 6.09\textdegree & 9.92\textdegree &  6.83 & 3.80 & 5.76 & 8.89 \\ \hdashline

Ours & \cellcolor[HTML]{FFFFBB} \textbf{135.71} &
\cellcolor[HTML]{FFFFBB} \textbf{31.21} &
\cellcolor[HTML]{FFFFBB} \textbf{68.63} &
\cellcolor[HTML]{FFFFBB} \textbf{151.49} &
\cellcolor[HTML]{FFFFBB} \textbf{5.41}\textdegree &
\cellcolor[HTML]{FFFFBB} \textbf{2.96}\textdegree &
\cellcolor[HTML]{FFFFBB} \textbf{4.45}\textdegree &
\cellcolor[HTML]{FFFFBB} \textbf{6.83}\textdegree &
\cellcolor[HTML]{FFFFBB} \textbf{5.24} &
\cellcolor[HTML]{FFFFBB} \textbf{3.32} &
\cellcolor[HTML]{FFFFBB} \textbf{4.57}  &
\cellcolor[HTML]{FFFFBB} \textbf{6.41} \\ \hline

\end{tabular}
}
\end{table*}

\begin{figure*}[t]
\centering
\includegraphics[width=\linewidth]{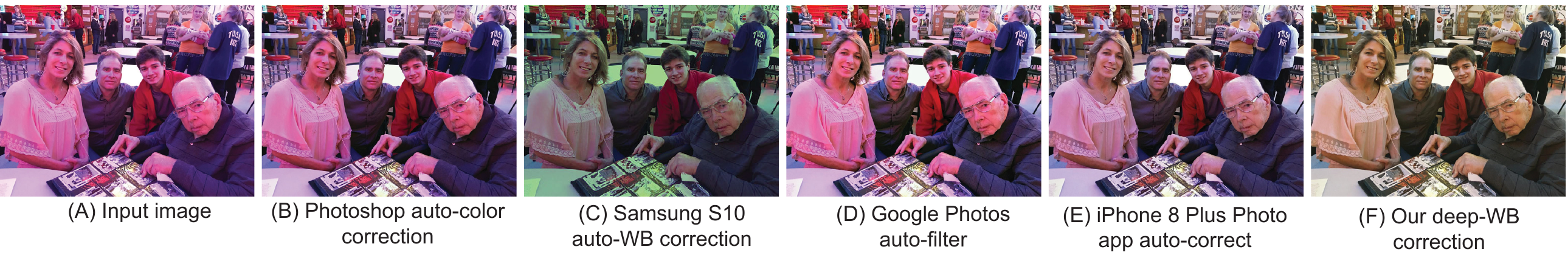}
\caption{Strong color casts due to WB errors are hard to correct. (A) Input image rendered with an incorrect WB setting. (B) Result of Photoshop auto-color correction. (C) Result of Samsung S10 auto-WB correction. (D) Result of Google Photos auto-filter. (E) Result of iPhone 8 Plus built-in Photo app auto-correction. (F) Our AWB result using the proposed deep-WB editing framework. Photo credit: \textit{OakleyOriginals} Flickr--CC BY 2.0.}
\label{fig:commercial}
\end{figure*}

Table\ \ref{Table0} reports the error between corrected images obtained by each method and the corresponding ground truth images.\ Table\ \ref{Table0} shows results on the Set 1-\textit{Test}, Set 2, and Cube+ dataset described earlier.  This represents a total of 34,169 unseen sRGB images by our DNN-model, each rendered with different camera models and WB settings. For the diagonal-correction results, we pre-processed each testing image by first applying the 2.2 gamma linearization \cite{anderson1996proposal, ebner2007color}, and then we applied the gamma encoding after correction. We have results that are on par with the state-of-the-art method \cite{afifi2019color} on the Set 1--\textit{Test}.  We achieve state-of-the-art results in all evaluation metrics for additional testing sets (Set 2 and Cube+).

\paragraph{WB manipulation} The goal of this task is to change the input image's colors to appear as they were rendered using a target WB setting. We compare our result with the most recent work in \cite{afifi2019ICCV} that proposed a KNN-WB emulator that mimics WB effects in the sRGB space. We used the same WB settings produced by the KNN-WB emulator. Specifically, we selected the following target WB settings: Incandescent ($2850K$), Fluorescent ($3800K$), Daylight ($5500K$), Cloudy ($6500K$), and Shade ($7500K$). As our decoders were trained to generate only Incandescent and Shade WB settings, we used Eq.\ \ref{eq6} to produce the other WB settings (i.e., Fluorescent, Daylight, and Cloudy WB settings).

Table\ \ref{Table1} shows the obtained results using our method and the KNN-WB emulator. Table\ \ref{Table1} demonstrates that our method outperforms the KNN-WB emulator \cite{afifi2019ICCV} over a total of 40,222 testing images captured with different camera models and WB settings using all evaluation metrics.

\subsection{Qualitative results}\label{subsec:qualitative}

In Fig.\ \ref{fig:qualitative_AWB} and Fig.\ \ref{fig:qualitative_WBEditing}, we provide a visual comparison of our results against the most recent work proposed for WB correction \cite{bianco2019quasi, afifi2019color} and WB manipulation \cite{afifi2019ICCV}, respectively. On top of each example, we show the $\bigtriangleup$E 2000 error between the result image and the corresponding ground truth image (i.e., rendered by the camera using the target setting). It is clear that our results have the lower $\bigtriangleup$E 2000 and are the most similar to the ground truth images.

Fig.\ \ref{fig:qualitative} shows additional examples of our results. As shown, our framework accepts input images with arbitrary WB settings and re-renders them with the target WB settings, including the AWB correction.

We tested our method with several images taken from the Internet to check its ability to generalize to images typically found online. Fig.\ \ref{fig:hard_case} and Fig.\ \ref{fig:commercial} show examples. As is shown, our method produces compelling results compared with other methods and commercial software packages for photo editing, even when input images have strong color casts.

\subsection{Comparison with a vanilla U-Net}

As explained earlier, our framework employs a single encoder to encode input images, while each decoder is responsible for producing a specific WB setting. Our architecture aims to model Eq.\ \ref{eq1} in the same way cameras would produce colors for different WB settings from the same raw-RGB captured image.

Intuitively, we can re-implement our framework using a multi-U-Net architecture \cite{unet}, such that each encoder/decoder model will be trained for a single target of the WB settings.

In Table\ \ref{Table2}, we provide a comparison between our proposed framework against vanilla U-Net models. We train our proposed architecture and three U-Net models (each U-Net model targets one of our WB settings) for 88,000 iterations. The results validate our design and make evident that our shared encoder not only reduces the required number of parameters but also gives better results.

\section{Conclusion}\label{sec:conclusion}

We have presented a deep learning framework for editing the WB of sRGB camera-rendered images.
Specifically, we have proposed a DNN architecture that uses a single encoder and multiple decoders, which are trained in an end-to-end manner. Our framework allows the direct correction of images captured with wrong WB settings. Additionally, our framework produces output images that allow users to manually adjust the sRGB image to appear as if it was rendered with a wide range of WB color temperatures.    Quantitative and qualitative results demonstrate the effectiveness of our framework against recent data-driven methods.

\begin{table}[]
\centering
\caption{Average of mean square error and $\boldsymbol{\bigtriangleup}$\textbf{E} 2000 \cite{sharma2005ciede2000} obtained by our framework and the traditional U-Net architecture \cite{unet}. Shown results on Set 2 of the Rendered WB dataset \cite{afifi2019color} for AWB and the rendered version of the Cube+ dataset \cite{banic2017unsupervised, afifi2019color} for WB manipulation. The top results are indicated with yellow and boldface.}
\label{Table2}
\scalebox{0.88}{
\begin{tabular}{|l|c|c|c|c|}
\hline
& \multicolumn{2}{|c|}{\cellcolor[HTML]{D4EBF2}\textbf{AWB \cite{afifi2019color}}} & \multicolumn{2}{|c|}{\cellcolor[HTML]{D4EBF2}\textbf{WB editing \cite{banic2017unsupervised, afifi2019color}}} \\ \cline{2-5}
\multicolumn{1}{|c|}{\multirow{-2}{*}{\textbf{Method}}} & \textbf{MSE} & \textbf{$\boldsymbol{\bigtriangleup}$\textbf{E} 2000} & \textbf{MSE} & \textbf{$\boldsymbol{\bigtriangleup}$\textbf{E} 2000} \\\hline

Multi-U-Net \cite{unet}  & 187.25
 & 6.23 & 234.77 &  6.87  \\\hdashline

Ours & \cellcolor[HTML]{FFFFBB}\textbf{124.47} &   \cellcolor[HTML]{FFFFBB}\textbf{4.99} & \cellcolor[HTML]{FFFFBB}\textbf{206.81} &  \cellcolor[HTML]{FFFFBB}\textbf{6.23} \\ \hline

\end{tabular}
}
\end{table}

\end{document}